\documentclass[letterpaper]{article} 
\usepackage{aaai2026} 
\usepackage{times} 
\usepackage{helvet} 
\usepackage{courier} 
\usepackage[hyphens]{url} 
\usepackage{graphicx} 
\usepackage{natbib}  
\usepackage{caption}  
\usepackage{amsmath} 
\usepackage{booktabs}
\usepackage{makecell}
\usepackage{amssymb}
\usepackage{subcaption}
\usepackage{float}

\title{ConfProBench: A Confidence Evaluation Benchmark for MLLM-Based Process Judges}

\author{
    Yue Zhou\textsuperscript{\rm 1} \quad
    Yi Chang\textsuperscript{\rm 1,2,3} \quad
    Yuan Wu\textsuperscript{\rm 1}\thanks{Corresponding author}
}
\affiliations {
    \textsuperscript{\rm 1}School of Artificial Intelligence, Jilin University\\
    \textsuperscript{\rm 2}Engineering Research Center of Knowledge-Driven Human-Machine Intelligence, MOE, China\\
    \textsuperscript{\rm 3}International Center of Future Science, Jilin University\\
    yuezhou24@mails.jlu.edu.cn, yichang@jlu.edu.cn, yuanwu@jlu.edu.cn \\
}

\begin{document}
\maketitle
\begin{abstract}
Reasoning is the critical capability of multimodal large language models (MLLMs) to solve complex multimodal tasks, and judging the correctness of reasoning steps is crucial to improving this capability. Recently, MLLM-based process judges (MPJs) have been widely used to judge the correctness of reasoning steps in multimodal reasoning tasks. Therefore, evaluating the capability of MPJs is crucial for identifying their limitations and guiding future improvements.
However, existing benchmarks for MPJs primarily focus on evaluating capabilities such as step correctness classification and reasoning process search, while overlooking a critical dimension: whether the confidence scores produced by MPJs at the step level are reliable. 
To fill this gap, we propose ConfProBench, the first comprehensive benchmark designed to systematically evaluate the reliability of step-level confidence scores generated by MPJs. 
This benchmark constructs three types of adversarially perturbed reasoning steps: Synonym Substitution, Syntactic Transformation, and Image Perturbation, to evaluate the robustness of MPJs' confidence under perturbations. Furthermore, we propose three novel evaluation metrics: Confidence Robustness Score (CRS), Confidence Sensitivity Score (CSS), and Confidence Calibration Score (CCS), which are designed to capture three complementary aspects of MPJs' confidence—robustness, sensitivity, and calibration. 
We evaluate 14 state-of-the-art MLLMs, including both proprietary and open-source models. Through extensive experiments, we reveal limitations in existing MPJs’ confidence performance and provide competitive baselines, thereby paving the way for future research in this field. 
\begin{links}
    \link{Code}{https://github.com/zy001122/confprobench}
\end{links}
\end{abstract}


\begin{figure}[t]
  \centering
  \includegraphics[width=1\columnwidth]{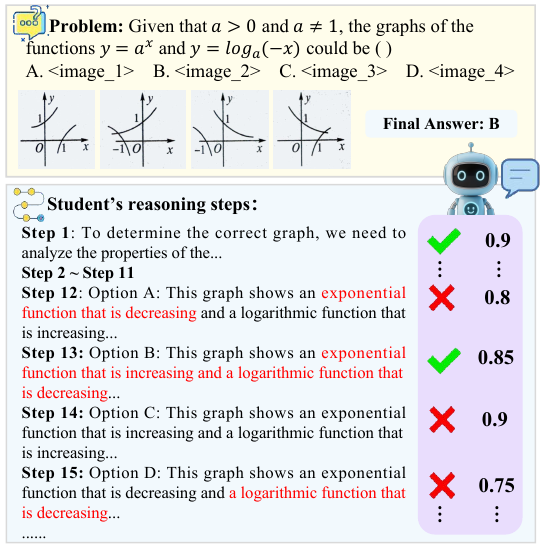}
  \caption{An example of the process judge task for MLLM-based process judges (MPJs), which perform binary classification of each reasoning step's correctness and provide associated confidence scores.}
  \label{fig:task_define}
\end{figure}

\section{Introduction}

\begin{table*}[htbp]
\centering
\setlength{\tabcolsep}{1.8mm} 
\label{tab:benchmark_comparison}
\begin{tabular}{lcccccc}
\toprule
\textbf{Benchmark} & \textbf{Multimodal} & \textbf{\makecell{Step\\Annotation}} & \textbf{\makecell{MPJ-specific \\ Confidence Metrics}} & \textbf{\makecell{Adversarial \\Perturbed Steps}} & \textbf{\makecell{Confidence \\Evaluation Paradigm}} &\\ 
\midrule
ProcessBench & No & Yes & No & No & No \\
PRMBench & No & Yes & Yes & No & No \\
VisualProcessBench & Yes & Yes & No & No & No \\
MPBench & Yes & Yes & No & No & No \\
ProJudgeBench & Yes & Yes & No & No & No \\
\midrule
ConfProBench (Ours) & Yes & Yes & Yes & Yes & Yes \\
\bottomrule
\end{tabular}
\caption{Comparison between related benchmarks with our ConfProBench.}
\end{table*}

Reasoning is a core capability of Multimodal Large Language Models (MLLMs) when tackling complex multimodal tasks~\cite{yan2024survey,shi2024math,li2025chemvlm,xiang2024atomthink}. 
Judging the correctness of each reasoning step is crucial for further enhancing this capability. As the reasoning chains generated by MLLMs become increasingly intricate, manually inspecting each intermediate step has become prohibitively costly. 
In response, recent studies have introduced MLLM-based Process Judges (MPJs) to assess step-by-step reasoning in multimodal tasks~\cite{pu2025judge,chen2024mllm,huang2024olympicarena,sun2024mm,zhang2024mathverse,jiang2025mme}. These MPJs analyze the reasoning process generated by MLLMs to identify potential flaws, improve interpretability, and facilitate targeted model improvements. 

However, this paradigm shift raises a fundamental question: Can we trust the judgments made by MPJs? To address this, existing benchmarks evaluate multiple aspects of MPJs, such as step correctness, error type identification, and answer aggregation \cite{ai2025projudge,xu2025mpbench,wang2025visualprm}. 
Nevertheless, they overlook an essential aspect: the reliability of the confidence scores produced by MPJs at the step level. 
Confidence not only reflects a model's self-assessed certainty but also directly affects controllability, reliability, and safety in downstream applications \cite{geng2023survey}. 
Under adversarial perturbations, robust and interpretable confidence scores are vital.

To fill this gap, we propose ConfProBench, the first benchmark specifically designed to systematically evaluate the confidence performance of MPJs. ConfProBench constructs perturbed variants of reasoning steps using three types of adversarial perturbations: Synonym Substitution, Syntactic Transformation, and Image Perturbation. These perturbations support the assessment of confidence robustness.

Furthermore, we introduce a comprehensive evaluation metric suite that includes three core components: Confidence Robustness Score (CRS), Confidence Sensitivity Score (CSS), and Confidence Calibration Score (CCS).
CRS measures the robustness of confidence under adversarial perturbations.
CSS measures the sensitivity of confidence scores to erroneous reasoning steps.
CCS evaluates the consistency between confidence scores and classification accuracy.

In summary, our main contributions are as follows:

\begin{itemize}
    \item We propose ConfProBench, the first benchmark dedicated to systematically evaluating the confidence performance of MPJs, and the first benchmark to assess confidence robustness and sensitivity.
    
    \item We construct three types of adversarial perturbation data to evaluate the robustness of MJPs' confidence. We further introduce the first comprehensive confidence evaluation suite for MPJs, consisting of three complementary metrics: CRS, CSS, and CCS, which assess robustness, sensitivity, and calibration.
    
    \item We conduct comprehensive experiments on 14 state-of-the-art MPJs, including both proprietary and open-source models. Through fine-grained analysis using the core metrics and their subcomponents, we reveal critical limitations in current models’ confidence performance and highlight directions for future improvement.
\end{itemize}

\section{Related Works}
\subsection{Confidence Evaluation and Estimation}
Confidence is the estimated probability that a model’s prediction matches the ground-truth label~\cite{guo2017calibration}. 
Assessing the confidence of large language models (LLMs) is essential for building reliable systems~\cite{geng2023survey}. 
Most studies focus on calibration, which measures how well predicted confidence aligns with actual prediction accuracy~\cite{zhao2024fact,geng2023survey}. 
Confidence estimation and evaluation are distinct: the former extracts signals from the model, while the latter assesses their trustworthiness and stability~\cite{geng2023survey}. 
Estimation methods include logit-based~\cite{duan2023shifting}, internal state-based~\cite{burns2022discovering}, consistency-based~\cite{manakul2023selfcheckgpt}, and verbalized approaches~\cite{xiong2023can}. 
Verbalized methods prompt LLMs to express confidence via natural language or numerical values, and are valued for their model-agnostic design and efficiency~\cite{geng2023survey,tian2023just,yang2024verbalized}. 
We adopt this approach by prompting MPJs to produce step-level verbalized confidence and evaluate its robustness, sensitivity, and calibration.

\subsection{Benchmarks for MLLM-based Process Judges}
In recent years, the process judgment capabilities of MLLMs have attracted increasing attention, and several related evaluation benchmarks have been proposed \cite{wang2025visualprm,xu2025mpbench,ai2025projudge}.
VisualProcessBench \cite{wang2025visualprm} provides human-annotated step-wise correctness labels to evaluate the ability of multimodal Process Reward Models (PRMs) to identify erroneous steps in multimodal reasoning tasks.
MPBench \cite{xu2025mpbench} aims to assess the performance of multimodal PRMs across three tasks: determining the correctness of each reasoning step (Step Correctness), selecting the optimal solution from multiple candidates (Answer Aggregation), and guiding the search of reasoning processes (Reasoning Process Search).
ProJudgeBench \cite{ai2025projudge} is a multimodal, multidisciplinary benchmark specifically designed to evaluate the fine-grained error detection, classification, and diagnosis capabilities of MPJs.
While existing benchmarks cover various aspects of multimodal process evaluation, they generally overlook a critical dimension: the confidence performance of process judges. To address this, we propose ConfProBench, a comprehensive benchmark for evaluating the confidence performance of MPJs.

\section{ConfProBench}
\subsection{Task Definition}
The multimodal process judging task in ConfProBench is framed as a binary classification problem. Our dataset contains two class labels: reasoning steps without errors are labeled as “correct” (1), while those with errors are labeled as “incorrect” (0).
Specifically, the MPJ is required to output the probability that a reasoning step belongs to the correct class, which is used for both classification and confidence scoring.

As illustrated in Figure~\ref{fig:task_define}, given a scientific problem $P$, its final answer $A$, and a step-by-step reasoning process $S = \{s_0, s_1, \cdots, s_{n-1}\}$ generated by a student model, the MPJ outputs a tuple $(l_i, p_i, e_i)$ for each reasoning step $s_i$.
Here, $l_i \in \{1, 0\}$ indicates whether $s_i$ is belong to the correct class ($l_i = 1$) or incorrect class($l_i = 0$); $p_i \in [0, 1]$ denotes the probability that $s_i$ belongs to the correct class; and $e_i$ represents the error type if $s_i$ is belongs to the incorrect class. 

The probability $p_i$ determines the predicted classification label and confidence score, while $l_i$ and $e_i$ assist in correcting potential inconsistencies in the result.

To obtain the binary step-level prediction, $p_i$ is converted into a correctness label $\hat{l}_i$ according to the following rule:
\begin{equation}
\hat{l}_i =
\begin{cases}
1, & \text{if } p_i \geq 0.5, \\
0, & \text{otherwise},
\end{cases}
\end{equation}

Based on $\hat{l}_i$ and $p_i$, the confidence score $c_i$ is defined as:
\begin{equation}
    c_i =
    \begin{cases}
        p_i, & \text{if } \hat{l}_i = 1, \\
        1 - p_i, & \text{if } \hat{l}_i = 0,
    \end{cases}
    \label{eq:conf_prob}
\end{equation}
$p_i$, $\hat{l}_i$, and $c_i$ are subsequently used to compute the proposed evaluation metrics.

\begin{figure}[t]
  \centering
  \includegraphics[width=1\columnwidth]{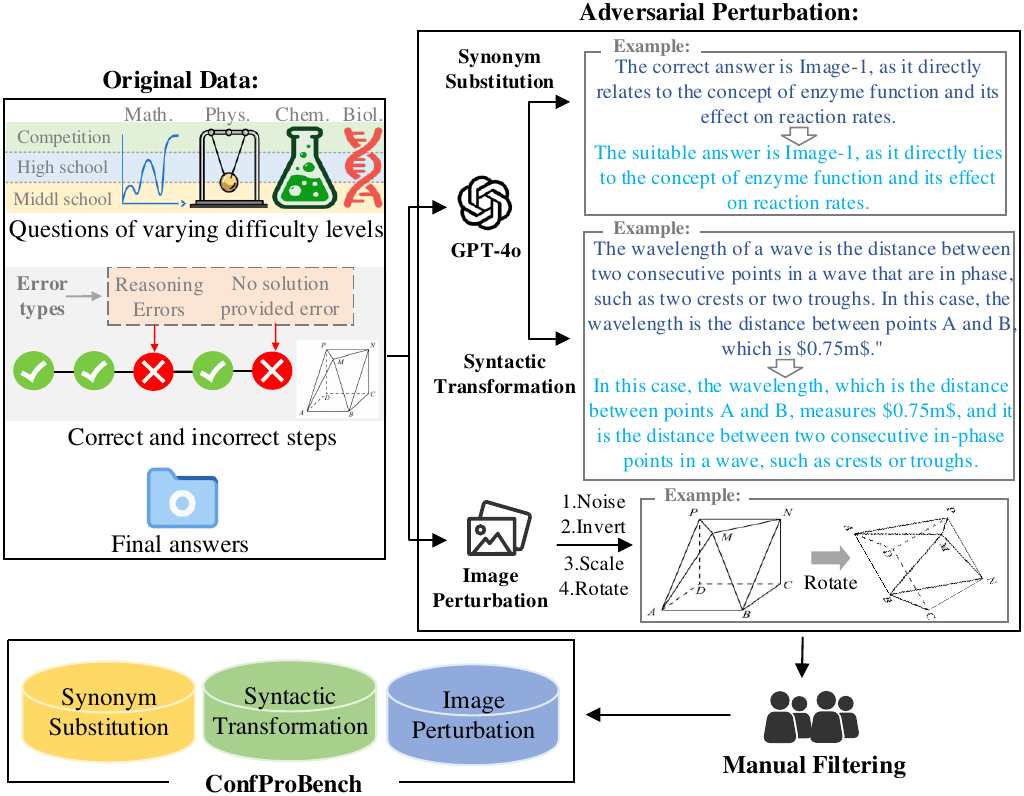}
  \caption{An overview of the data construction process for ConfProBench.}
  \label{fig:data construct}
\end{figure}

\subsection{Dataset Construction}
\paragraph{Meta Data Extraction.}
We construct our benchmark based on ProJudgeBench \cite{ai2025projudge} by sampling 1,200 problems spanning three difficulty levels (Middle School, High School and Competition), four scientific disciplines (Math, Physics, Chemistry, Biology), three modality types (Single Image, Multi Images, Pure Text), and seven types of reasoning errors (Numerical Calculation Error, Reasoning Error, Symbolic Calculation Error, Knowledge Error, Visual Interpretation Error, Question Understanding Error, and No Solution Provided).
The resulting dataset maintains a balanced distribution across difficulty levels and scientific disciplines, offering a robust foundation for a comprehensive evaluation of MPJs' confidence performance.
Please refer to Appendix~A for detailed statistics of ConfProBench.


\paragraph{Adversarial Perturbation Construction.}
We design three types of perturbations: (1) Synonym Substitution, (2) Syntactic Transformation, and (3) Image Perturbation. Note that image perturbations are only applicable to samples with the Single Image or Multi Images modality. We divide the 1,200 scientific problems into three equal subsets, with each subset applying only one perturbation type.

\textbf{Synonym Substitution}: We prompt GPT-4o to generate five distinct synonym-substituted versions for each reasoning step and randomly select one.
In each version, at least one non-technical term, such as mathematical symbols, scientific terminology, programming syntax, technical jargon, or domain-specific abbreviations, is replaced with a semantically equivalent synonym. As many such terms as possible are substituted while ensuring grammatical correctness and semantic consistency. 

\textbf{Syntactic Transformation}: We prompt GPT-4o to generate five distinct syntactic transformation versions for each step that preserve the original semantic information while exhibiting distinct syntactic structures, and randomly select one.
Each syntactic transformation version strictly applies one of the following six predefined syntactic transformations: (1) voice alternation (active to passive), (2) adverbial position adjustment, (3) clause order or structural variation, (4) phrase simplification or expansion, (5) inversion or emphasis construction, and (6) transformation of conditional, purposive, or resultative constructions.

\textbf{Image Perturbation}: We apply image-level perturbations to the image inputs of multimodal scientific problems. Specifically, one image transformation is randomly selected from the following set of operations: scaling, rotation, Gaussian noise injection, or color inversion. These transformations are designed to modify the low-level visual features of the input while preserving its semantic information.

Examples of each perturbation type are shown in Figure~\ref{fig:data construct}.

\paragraph{Data Quality Control.}
We conducted comprehensive manual verification of all adversarial perturbation results to ensure their quality and validity. 

Each reasoning step with synonym substitution was examined to ensure that: (1) at least one non-technical term was replaced; (2) the original syntactic structure and semantic information were preserved; (3) technical terms and domain-specific vocabulary remained unchanged; (4) numerical values and mathematical expressions were not modified; and (5) the rewritten step was grammatically correct and fluent.
Each syntactically transformed reasoning step was reviewed to ensure that: (1) no mathematical derivations, intermediate steps, or key expressions were omitted; (2) all numerical and symbolic content remained intact; (3) the sentence maintained its original meaning; and (4) the target structural transformation was appropriately applied.
For image perturbations, we examined each transformed image to ensure that the applied modifications did not introduce semantic information drift or obscure essential visual information.
If a perturbed result failed to meet these criteria, we re-applied the corresponding perturbation procedure to the same reasoning step until a valid adversarial variant was obtained.

\subsection{Evaluation Metrics}
\begin{figure}[t]
  \centering
  \includegraphics[width=1\columnwidth]{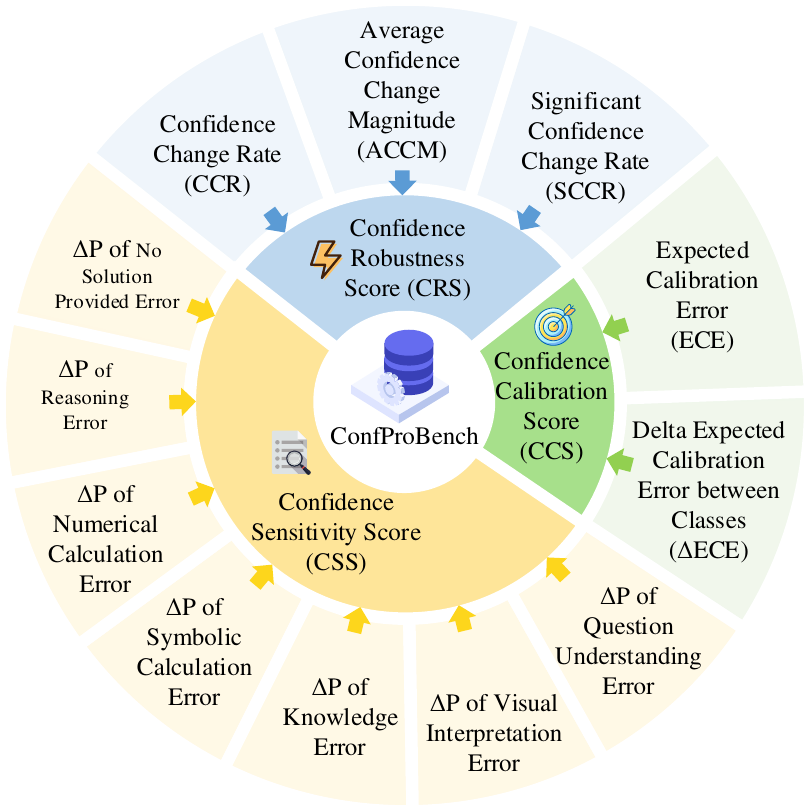}
  \caption{An overview of the proposed evaluation metric suite, which consists of three core metrics: Confidence Robustness Score (CRS), Confidence Sensitivity Score (CSS), and Confidence Calibration Score (CCS). Each core metric is composed of a set of sub-metrics.}
  \label{fig:metric}
\end{figure}

To comprehensively evaluate the reliability of confidence scores produced by MPJs, we introduce a multi-dimensional suite of evaluation metrics, as illustrated in Figure~\ref{fig:metric}. This metric suite is designed to capture three complementary aspects of confidence performance: robustness, sensitivity, and calibration. These three metrics form a comprehensive framework to assess whether an MPJ can reliably express the uncertainty of its predictions, which is an essential capability for trustworthy MPJs.

\paragraph{Confidence Robustness Score (CRS).}
We define the Confidence Robustness Score (CRS) to measure the robustness of confidence under designed adversarial perturbations, including Synonym Substitution, Syntactic Transformation, and Image Perturbation. Since these perturbations preserve the semantic consistency of the reasoning steps, an ideal process judge should maintain consistent confidence scores across both perturbed and unperturbed inputs.

CRS integrates three sub-metrics to quantify confidence robustness. Let \( c_i \) represent the original confidence score, and \( c_i' \) represent the confidence score after perturbation. For each pair of original confidence score and post-perturbation confidence score, we compute the following sub-metrics:

\textbf{Confidence Change Rate (CCR)}: The proportion of reasoning steps in which the confidence scores change after perturbation. Specifically, if the absolute difference in confidence exceeds a small threshold $\epsilon$ (set to 0.01), we consider the confidence to have changed. CCR is defined as:
\begin{equation}
\mathrm{CCR} = \frac{1}{N} \sum_{i=1}^{N} \mathbb{I} \left( |c_i - c_i'| > \epsilon \right),
\end{equation}
Where \( N \) is the total number of reasoning steps, and \( \mathbb{I}(\cdot) \) is the indicator function, which is used to check if a condition is met. It returns 1 if the condition is true, and 0 if it is false. 
A lower CCR indicates greater robustness of confidence.

\textbf{Average Confidence Change Magnitude (ACCM)}: The average magnitude of confidence change across all steps where the change exceeds the small threshold $\epsilon$ (set to 0.01). Specifically, we define:
\begin{equation}
\begin{aligned}
\text{ACCM} &= \frac{1}{|S|} \sum_{i \in S} |c_i - c_i'|, \\
\text{where } S &= \left\{i \mid |c_i - c_i'| > \epsilon \right\},
\end{aligned}
\end{equation}
A smaller ACCM indicates greater robustness of confidence.

\textbf{Significant Confidence Change Rate (SCCR)}: It refers to the proportion of reasoning steps where the confidence score changes beyond a predefined threshold \(\delta\). We refer to this threshold as the significant threshold, which is set to \(0.2\) in our experiments. This parameter can be adjusted according to different application needs. The formal definition of SCCR is as follows:
\begin{equation}
\text{SCCR} = \frac{1}{N} \sum_{i=1}^{N} \mathbb{I} \left( |c_i - c_i'| > \delta \right),
\end{equation}
A lower SCCR indicates greater robustness of confidence.

We combine the three sub-metrics above to define the CRS as follows:
\begin{equation}
\begin{aligned}
\text{CRS} =\ & w_1 \cdot (1 - \text{CCR})  + w_2 \cdot (1 - s \cdot \text{ACCM}) \\
              & + w_3 \cdot (1 - s \cdot \text{SCCR}),
\end{aligned}
\end{equation}
where $w_1$, $w_2$, and $w_3$ are the weights of the three sub-metrics, and $s$ is a scaling factor. Since the values of ACCM and SCCR are typically much smaller than that of CCR, we introduce $s$ to magnify ACCM and SCCR by a factor of $s$ (set to 5), preventing their influence on CRS from being diminished. All sub-metrics are defined such that lower values indicate better performance. Accordingly, each sub-metric is subtracted from 1 to ensure that a higher CRS corresponds to greater confidence robustness. 

In our experiments, we set $w_1 = 0.4$, $w_2 = 0.4$, $w_3 = 0.2$. This reflects our greater emphasis on CCR and ACCM. The weights are adjustable for different scenarios.
Under this configuration, CRS ranges from $[-2.4, 1]$, where a score of 1 denotes perfect confidence robustness, meaning that the MPJ exhibits no confidence change under adversarial perturbations.
We set $s = 5$ based on extensive experimental results. This ensures that each sub-metric contributes effectively.

\begin{table}[t]
\centering
\begin{tabular}{lcccc}
\toprule
\textbf{Model} & \textbf{CRS$\uparrow$} & \textbf{CSS$\uparrow$} & \textbf{CCS$\uparrow$} & \textbf{Avg.$\uparrow$} \\ 
\midrule
\multicolumn{5}{c}{\textbf{Open-source MLLMs}} \\ \midrule
InternVL3-8B           & 77.41 & 11.55 & 25.97   & 38.31 \\
InternVL3-14B          & 50.78 & \underline{21.19} & \textbf{46.75}   & 39.57 \\
InternVL3-38B          & 49.92 & \textbf{30.62} & \underline{44.49}   & \underline{41.68} \\
MiniCPM-V-2 6          & 68.05 & 6.60  & -47.95  & 8.90  \\
Qwen2.5-VL-3B          & 74.71 & 3.15  & 2.73    & 26.86 \\
Qwen2.5-VL-7B          & 71.19 & 10.38 & 15.80   & 32.46 \\
Qwen2.5-VL-32B         & \textbf{81.06} & 15.93 & 41.60   & \textbf{46.20} \\
Qwen2.5-VL-72B         & \underline{77.45} & 19.93 & 25.30   & 40.89 \\
QVQ                   & 74.17 & 12.60 & 30.69   & 39.15 \\
\midrule
\multicolumn{5}{c}{\textbf{Proprietary MLLMs}} \\ \midrule
GPT-4o                 & 57.37 & 30.71 & \textbf{62.00}   & \underline{50.03} \\
GPT-4o-Mini            & \underline{65.58} & 13.03 & 47.73            & 42.11 \\
GPT-4.1                & \textbf{73.62} & 38.51 & 37.65   & 49.93 \\
Gemini-2.5-flash       & 63.08 & \textbf{48.29} & 48.62   & \textbf{53.33} \\
\makecell[l]{Gemini-2.5-flash\\-nothinking} & 51.20 & \underline{42.13} & \underline{51.55} & 48.29 \\  
\bottomrule
\end{tabular}
\caption{The main results across different MLLM-based Process
Judges (MPJs) on ConfProBench. The best performance for each metric is shown in bold, while the second-best is underlined.}
\label{tab:main_table}
\end{table}

\begin{table}[t]  
\centering
\begin{tabular}{lccc}
\toprule
\textbf{Model} & \textbf{\makecell{CCR$\downarrow$}}
 & \textbf{\makecell{ACCM$\downarrow$}} & \textbf{\makecell{SCCR$\downarrow$}} \\ 
\midrule
\multicolumn{4}{c}{\textbf{Open-source MLLMs}} \\ \midrule
InternVL3-8B       & 21.47 & \underline{6.56} & 0.89 \\
InternVL3-14B      & 61.57 & 9.35 & 5.89 \\
InternVL3-38B      & 61.71 & 9.26 & 6.87 \\
MiniCPM-V-2 6      & 22.49 & 10.68 & 1.60 \\
Qwen2.5-VL-3B      & \textbf{15.78} & 8.65 & 1.69 \\
Qwen2.5-VL-7B      & 24.97 & 8.31 & 2.19 \\
Qwen2.5-VL-32B     & \underline{15.83} & \textbf{6.29} & \textbf{0.04} \\
Qwen2.5-VL-72B     & 21.81 & 6.63 & \underline{0.58} \\
QVQ                & 28.68 & 6.74 & 0.86 \\
\midrule

\multicolumn{4}{c}{\textbf{Proprietary MLLMs}} \\ \midrule
GPT-4o                     & \textbf{21.82} & 13.46 & 6.98 \\
GPT-4o-Mini                & \underline{27.68} & 9.55  & \underline{4.24} \\
GPT-4.1                    & 34.96 & \textbf{5.47}  & \textbf{1.46} \\
Gemini-2.5-flash           & 38.56 & \underline{8.17}  & 5.15 \\
\makecell[l]{Gemini-2.5-flash\\-nothinking} & 40.31 & 11.77 & 9.12 \\

\bottomrule
\end{tabular}
\caption{The results of the sub-metrics that constitute the Confidence Robustness Score (CRS). The best performance for each metric is shown in bold, while the second-best is underlined.}
\label{tab:CRS_decomposition}
\end{table}
\paragraph{Confidence Sensitivity Score (CSS).}
We propose Confidence Sensitivity Score (CSS), a novel metric that quantifies how sensitively confidence scores respond to reasoning errors. 

For each error type $t \in \mathcal{T}$, let $\overline{p}_t$ denote the average value of $p_i$ over all steps labeled with the ground-truth error type $t$, and let $\overline{p}_{\text{correct}}$ denote the average $p_i$ over all steps labeled as ground-truth correct. We then define $\Delta p_t$ as the difference between $\overline{p}_{\text{correct}}$ and $\overline{p}_t$, as follows:
\begin{equation}
\Delta p_t = \overline{p}_{\text{correct}} - \overline{p}_t,
\end{equation}
A larger \(\Delta p_t\) indicates that \(p_i\) significantly decreases when encountering an error of type \( t \), showing that \(p_i\) is sensitive to this type of error. Conversely, a smaller or even negative \(\Delta p_t\) suggests that \(p_i\) has weak or no ability to recognize this error type. 
Since \( c_i \) is derived from \(p_i\) through a simple linear transformation, the sensitivity of \(p_i\) to reasoning errors directly reflects the model confidence's sensitivity.

To assess the overall confidence sensitivity, we define CSS as the average of $\Delta p_t$ across all error types:
\begin{equation}
\text{CSS} = \frac{1}{|\mathcal{T}|} \sum_{t \in \mathcal{T}} \Delta p_t,
\end{equation}
where \(\mathcal{T}\) is the set of all non-empty error types in the dataset. Since each \(\Delta p_t\) lies in the range \([-1, 1]\), the CSS also falls within this interval.

\paragraph{Confidence Calibration Score (CCS).}
Confidence Calibration Score (CCS) evaluates the consistency between the confidence score and the actual accuracy of predictions . It incorporates two aspects of calibration errors: the overall Expected Calibration Error (ECE) \cite{guo2017calibration}, and the gap in ECE between classes, denoted as $\Delta \mathrm{ECE}_{\text{cls}}$.

The ECE is defined as:
\begin{equation}
\text{ECE} = \sum_{m=1}^{M} \frac{|B_m|}{n} \cdot \left| \text{acc}(B_m) - \text{conf}(B_m) \right|,
\end{equation}
where $B_m$ is the $m$-th bin obtained by equally dividing the confidence range into $M$ intervals, $|B_m|$ denotes the number of samples in bin $B_m$, and $n$ is the total number of samples. $\text{acc}(B_m)$ and $\text{conf}(B_m)$ represent the average accuracy and average confidence of samples within bin $B_m$, respectively.

To better capture class-specific calibration performance, we compute the ECE separately for the correct and incorrect categories of reasoning steps, denoted as: 
\(\mathrm{ECE}_{\text{correct}}\) and \(\mathrm{ECE}_{\text{incorrect}}\). The class-wise calibration gap is then defined as
\begin{equation}
\Delta \mathrm{ECE}_{\text{cls}} = \left| \mathrm{ECE}_{\text{correct}} - \mathrm{ECE}_{\text{incorrect}} \right|,
\end{equation}

A smaller $\Delta \mathrm{ECE}_{\text{cls}}$ indicates more balanced confidence calibration performance across different classes.

Combining ECE and $\Delta \mathrm{ECE}_{\text{cls}}$, we define CCS as follows:
\begin{equation}
\text{CCS} = 0.5 \cdot (1 - s \cdot \text{ECE}) + 0.5 \cdot (1 - \Delta \mathrm{ECE}_{\text{cls}}),
\end{equation}
Where $s$ (set to 5) is a scaling factor. Since the value of ECE is typically smaller than that of $\Delta \mathrm{ECE}_{\text{cls}}$, this factor ensures that changes in ECE have a meaningful impact on the CCS.

Since $\text{ECE}, \Delta \mathrm{ECE}_{\text{cls}} \in [0, 1]$, the theoretical range of CCS is $[-2, 1]$. A higher CCS reflects stronger confidence calibration.

\begin{table*}[t] 
\centering
\begin{tabular}{lccccccc}
\toprule
\textbf{Model} & $\Delta p_{\text{NSPE}} \uparrow$ & $\Delta p_{\text{RE}} \uparrow$ & $\Delta p_{\text{NCE}} \uparrow$ & $\Delta p_{\text{SCE}} \uparrow$ & $\Delta p_{\text{KE}} \uparrow$ & $\Delta p_{\text{VIE}} \uparrow$ & $\Delta p_{\text{QUE}} \uparrow$ \\

\midrule
\multicolumn{8}{c}{\textbf{Open-source MLLMs}} \\ \midrule
InternVL3-8B       & 8.80 & 13.48 & 4.40 & 7.59 & 18.86 & 6.36 & \underline{21.36} \\
InternVL3-14B      & 10.24 & \textbf{28.28} & \underline{28.35} & 19.58 & 22.79 & \underline{15.26} & \textbf{23.85} \\
InternVL3-38B      & \textbf{60.46} & 35.14 & \textbf{31.79} & \textbf{24.81} & \textbf{32.84} & \textbf{17.94} & 11.38 \\
MiniCPM-V-2 6      & 19.95 & 10.01 & 13.23 & 15.46 & 2.03 & 7.12 & -21.62 \\
Qwen2.5-VL-3B      & 12.40 & 1.67 & 0.02 & 2.12 & 8.35 & 1.73 & -4.22 \\
Qwen2.5-VL-7B      & 18.96 & 10.66 & 5.32 & 5.78 & 15.48 & 8.59 & 7.87 \\
Qwen2.5-VL-32B     & 11.23 & 25.60 & 18.59 & 17.87 & 16.79 & 7.76 & 13.64 \\
Qwen2.5-VL-72B     & 17.18 & \underline{28.24} & 21.43 & \underline{20.16} & \underline{23.90} & 9.10 & 19.47 \\
QVQ                & \underline{24.98} & 12.91 & 6.77 & 14.28 & 14.01 & 6.81 & 8.49 \\
\midrule
\multicolumn{8}{c}{\textbf{Proprietary MLLMs}} \\ \midrule
GPT-4o             & \textbf{48.34} & 35.12 & 34.73 & 32.53 & 28.67 & 21.53 & 14.01\\
GPT-4o-Mini        & 7.22 & 18.28 & 13.67 & 23.08 & 11.83 & 7.35 & 9.81\\
GPT-4.1            & 2.38 & \underline{51.77} & \textbf{56.23} & 45.68 & \underline{45.15} & \underline{40.83} & 27.49\\
Gemini-2.5-flash   & \underline{27.60} & \textbf{54.03} & \underline{54.08} & \textbf{53.99} & \textbf{53.89} & \textbf{49.48} & \underline{44.94}\\
\makecell{Gemini-2.5-flash-nothinking} & 23.17 & 49.34 & 41.57 & \underline{46.61} & 41.85 & 40.57 & \textbf{51.77}\\
\bottomrule
\end{tabular}
\caption{The results of the sub-metrics that constitute the Confidence Sensitivity Score (CSS). The best performance for each metric is shown in bold, while the second-best is underlined. NCE denotes Numerical Calculation Error, RE denotes Reasoning Error, SCE denotes Symbolic Calculation Error, KE denotes Knowledge Error, VIE denotes Visual Interpretation Error, QUE denotes Question Understanding Error, and NSPE denotes No Solution Provided Error.}
\label{tab:CSS_decomposition}
\end{table*}

\section{Experiments}
\subsection{Experimental Settings}
To provide a comprehensive evaluation on ConfProBench, we assess both proprietary and open-source MPJs.
The proprietary MPJs include GPT-4o \cite{openai2024gpt4o}, GPT-4o-Mini \cite{openai2024gpt4omini}, GPT-4.1 \cite{openai2024gpt4_1}, Gemini-2.5-flash (Dynamic thinking) \cite{deepmind2025gemini25flash}, and Gemini-2.5-flash-nothinking \cite{google2025gemini25flash_nothinking}. The open-source MPJs span a variety of architectures and parameter scales, including InternVL3 (8B, 14B, 38B) \cite{zhu2025internvl3}, Qwen2.5-VL (3B, 7B, 32B, 72B) \cite{bai2025qwen2}, MiniCPM-V-2\_6 (8B) \cite{yao2024minicpm}, and QVQ (72B) \cite{qwen2024qvq72b}.

All MPJs use a unified prompt template, with detailed prompt designs provided in Appendix~B. All metric values are multiplied by 100\% for presentation in the tables.

\subsection{Results and Analysis}
The primary experimental results for the three core metrics CRS, CSS, and CCS are presented in Table~\ref{tab:main_table}. To enable more fine-grained analysis, the results of the sub-metrics that constitute these core metrics are reported separately in Tables~\ref{tab:CRS_decomposition}–\ref{tab:CCS_decomposition}. 

\paragraph{Confidence Robustness Analysis.}
As shown in Table~\ref{tab:main_table}, GPT-4.1 achieves the highest CRS (73.62) among all proprietary MPJs. However, several open-source MPJs, including InternVL3-8B (77.41), Qwen2.5-VL-3B (74.71), Qwen2.5-VL-32B (81.06), Qwen2.5-VL-72B (77.45), and QVQ (74.17), outperform GPT-4.1 on CRS.
This observation indicates that under adversarial perturbations, the confidence robustness of proprietary MPJs is not always superior to that of open-source MPJs, highlighting the effectiveness of the CRS metric in uncovering weaknesses in confidence robustness.
Nonetheless, the CRS scores of the best-performing MPJs still fall short of the theoretical maximum, suggesting that there remains substantial room for improvement in current MPJs.

As shown in Table~\ref{tab:CRS_decomposition}, the three sub-metrics of CRS provide a more fine-grained explanation of the differences in confidence robustness among MPJs. For example, Qwen2.5-VL-32B shows low CCR (15.83), ACCM (6.29), and SCCR (0.04), indicating that confidence changes are infrequent, mild, and rarely exceed the significance threshold under adversarial perturbations. 
Therefore, its CRS is the highest among all MPJs, reflecting the best confidence robustness. In contrast, InternVL3-38B shows much higher CCR (61.71), ACCM (9.26), and SCCR (6.87), indicating frequent, large, and significant confidence changes under perturbations. As a result, it has the lowest CRS and weakest confidence robustness among all MPJs.

\begin{table}[htbp]  
\centering
\begin{tabular}{lcccc}
\toprule
\textbf{Model} & \textbf{\makecell{ECE\\(C.)$\downarrow$}} & \textbf{\makecell{ECE\\(I.)$\downarrow$}} & \textbf{$\Delta$ECE$\downarrow$} & \textbf{ECE$\downarrow$} \\ 
\midrule
\multicolumn{5}{c}{\textbf{Open-source MLLMs}} \\ \midrule
InternVL3-8B       &8.86 &90.18 &81.32 &13.35 \\
InternVL3-14B      &10.71 &85.05 &\underline{74.34} &\textbf{6.43} \\
InternVL3-38B      &8.80 &\underline{84.82} &76.03 &\underline{7.00} \\
MiniCPM-V-2 6      &16.51 &\textbf{84.61} &\textbf{68.09} &45.16 \\
Qwen2.5-VL-3B      &9.24 &90.50 &81.26 &22.66 \\
Qwen2.5-VL-7B      &9.24 &88.85 &79.62 &17.76 \\
Qwen2.5-VL-32B     &9.41 &88.88 &79.47 &7.47 \\
Qwen2.5-VL-72B     &\textbf{4.37} &92.16 &87.78 &12.32 \\ 
QVQ                &\underline{8.25} &89.72 &81.47 &11.43 \\
\midrule
\multicolumn{5}{c}{\textbf{Proprietary MLLMs}} \\ \midrule
GPT-4o             &10.54 &\textbf{76.93} &\textbf{66.39} &\textbf{1.92} \\
GPT-4o-Mini        &10.32 &83.33 &73.01 &6.31 \\
GPT-4.1            &\textbf{3.00} &89.81 &86.81 &7.58 \\
Gemini-2.5-flash   &\underline{6.43} &87.56 &81.13 &\underline{4.32} \\
\makecell[l]{Gemini-2.5-flash\\-nothinking} &9.06 &\underline{82.02} &\underline{72.97} &4.79 \\
\bottomrule
\end{tabular}
\caption{The results of the sub-metrics that constitute the Confidence Calibration Score (CCS). C. indicates the correct class, and I. indicates the incorrect class. The best performance for each metric is shown in bold, while the second-best is underlined.}
\label{tab:CCS_decomposition}
\end{table}
\paragraph{Confidence Sensitivity Analysis.}  
As shown in Table~\ref{tab:main_table}, proprietary MPJs outperform open-source MPJs in CSS, with Gemini-2.5-flash achieving the highest score (48.29), followed by its no-thinking variant (42.13). However, there remains substantial room for improvement compared to the theoretical upper bound of confidence sensitivity. 
Table~\ref{tab:CSS_decomposition} shows that proprietary MPJs achieve higher $\Delta p$ across most error types, indicating stronger confidence sensitivity. Some open-source MPJs, such as Qwen2.5-VL-3B ($-4.22$) and MiniCPM-V-2 6 ($-21.62$), exhibit negative $\Delta p$ on QUE (Question Understanding Error), suggesting that their confidence lacks sensitivity to QUE. In such cases, the confidence scores are unreliable, as they fail to distinguish between reasoning steps that belong to the correct and incorrect classes.

\paragraph{Confidence Calibration Analysis.}
As shown in Table~\ref{tab:main_table}, proprietary MPJs significantly outperform open-source MPJs in CCS. Among them, GPT-4o achieves the highest CCS score of 62.00, indicating substantially stronger confidence calibration performance than other MPJs. However, this is still far from the theoretical upper bound, suggesting ample room for further improvement. In contrast, open-source MPJs such as MiniCPM-V-2 6 and Qwen2.5-VL-3B perform poorly, with MiniCPM-V-2 6 exhibiting a notably high ECE of 45.16, resulting in a negative CCS score of \(-47.95\).
Furthermore, as shown in Table \ref{tab:CCS_decomposition}, it can be observed that across all MPJs, \(\text{ECE}_{\text{correct}}\) is consistently much lower than \(\text{ECE}_{\text{incorrect}}\), resulting in relatively large \(\Delta\text{ECE}\) values. This indicates an imbalance in confidence calibration across classes. Therefore, the calibration performance on erroneous reasoning steps remains unsatisfactory and calls for urgent improvement.

\paragraph{Average Score Comparison.}
As shown in Table \ref{tab:main_table}, Gemini-2.5-flash ranks first in the average score across the three confidence metrics (CRS, CSS, CCS), achieving 53.33, followed by GPT-4o and GPT-4.1. These three MPJs demonstrate relatively balanced performance across three metrics. Most open-source MPJs score between 30 and 40, with MiniCPM-V-2 6 being the lowest at 8.90, primarily due to its negative CCS (indicating poor calibration performance).
Among open-source MPJs, the InternVL series consistently outperforms the Qwen2.5-VL series, and its scores show an upward trend with increasing model size. Notably, InternVL3-38B performs best with a score of 41.68. Although the Qwen2.5-VL series improves steadily from 3B to 32B, its performance drops at 72B, suggesting that continued increasing model size alone is insufficient to ensure improved confidence performance.

In addition to overall CRS scores, we report CRS under three types of adversarial perturbations. MPJ shows the lowest robustness under syntactic transformations. Appendix~C provides detailed results and analysis. We also report CRS, CSS, and CCS across different difficulty levels, subjects, and modalities, with detailed results and analysis in Appendix~D.

\paragraph{Impact of Model Scale on Confidence Performance.}
As shown in Table~\ref{tab:main_table}, model scale exhibits varying effects on different aspects of confidence performance.
Specifically, no clear positive correlation is observed between model size and confidence robustness. For example, within the InternVL3 series, CRS consistently decreases as model size increases from 8B to 38B.
In contrast, confidence sensitivity generally improves with scale. For instance, in the Qwen2.5-VL series, CSS rises from 3.15 (3B) to 19.93 (72B), indicating enhanced confidence sensitivity.
As for calibration, larger models tend to perform better. For example, Qwen2.5-VL’s CCS increases from 2.73 (3B) to 41.60 (32B), but drops again at 72B, suggesting that increasing model size alone does not ensure better calibration.

\paragraph{Impact of Thinking Mode on Confidence Performance.}
Table~\ref{tab:main_table} presents the core metric results for Gemini-2.5-flash and its no-thinking variant.
Results show that enabling the thinking process enhances confidence robustness under input perturbations, as evidenced by a higher CRS. 
Additionally, Gemini-2.5-flash exhibits a 6.16-point improvement in CSS, suggesting that the thinking process enhances the model’s sensitivity to erroneous reasoning steps. However, its CCS is lower than that of the no-thinking variant, indicating that the thinking process does not necessarily improve confidence calibration quality.

\section{Conclusion}
We present ConfProBench, the first benchmark for evaluating the reliability of step-level confidence scores produced by MPJs.
It introduces three types of adversarial perturbations to assess the robustness of MPJs' confidence under input variations.
Furthermore, it proposes a comprehensive evaluation suite comprising three complementary metrics: CRS, CSS, and CCS, which measure the robustness, sensitivity, and calibration of MPJs' confidence.
Extensive experiments reveal key limitations in current MPJs’ confidence performance and establish strong baselines, paving the way for future research in this area.

Beyond these contributions, we suggest two future directions.
First, conducting human confidence annotations and introducing new consistency metrics to assess the alignment between MPJ confidence and expert judgments.
Second, extending ConfProBench to encompass safety-critical scenarios where highly reliable confidence estimation is essential.

\bibliography{aaai2026}

\clearpage
\appendix

\section{Detailed Statistics of ConfProBench}
The detailed statistics of ConfProBench are summarized in Table~\ref{tab:Statistics of ConfProBench}.

\section{Prompt for Adversarial Perturbations Generation and Process Judging}
The prompt used to generate reasoning steps with syntactic transformation perturbations is shown in Table~\ref{tab:syntactic_transformation_prompt}.
The prompt used to generate reasoning steps with synonym substitution perturbations is shown in Table~\ref{tab:synonym_substitution_prompt}.
The prompt used for the multimodal process judging task is shown in Table~\ref{tab:process_judging_prompt}.

\section{Confidence Robustness Across Perturbation Types}
\begin{figure}[t]
  \centering
  \includegraphics[width=0.95\linewidth]{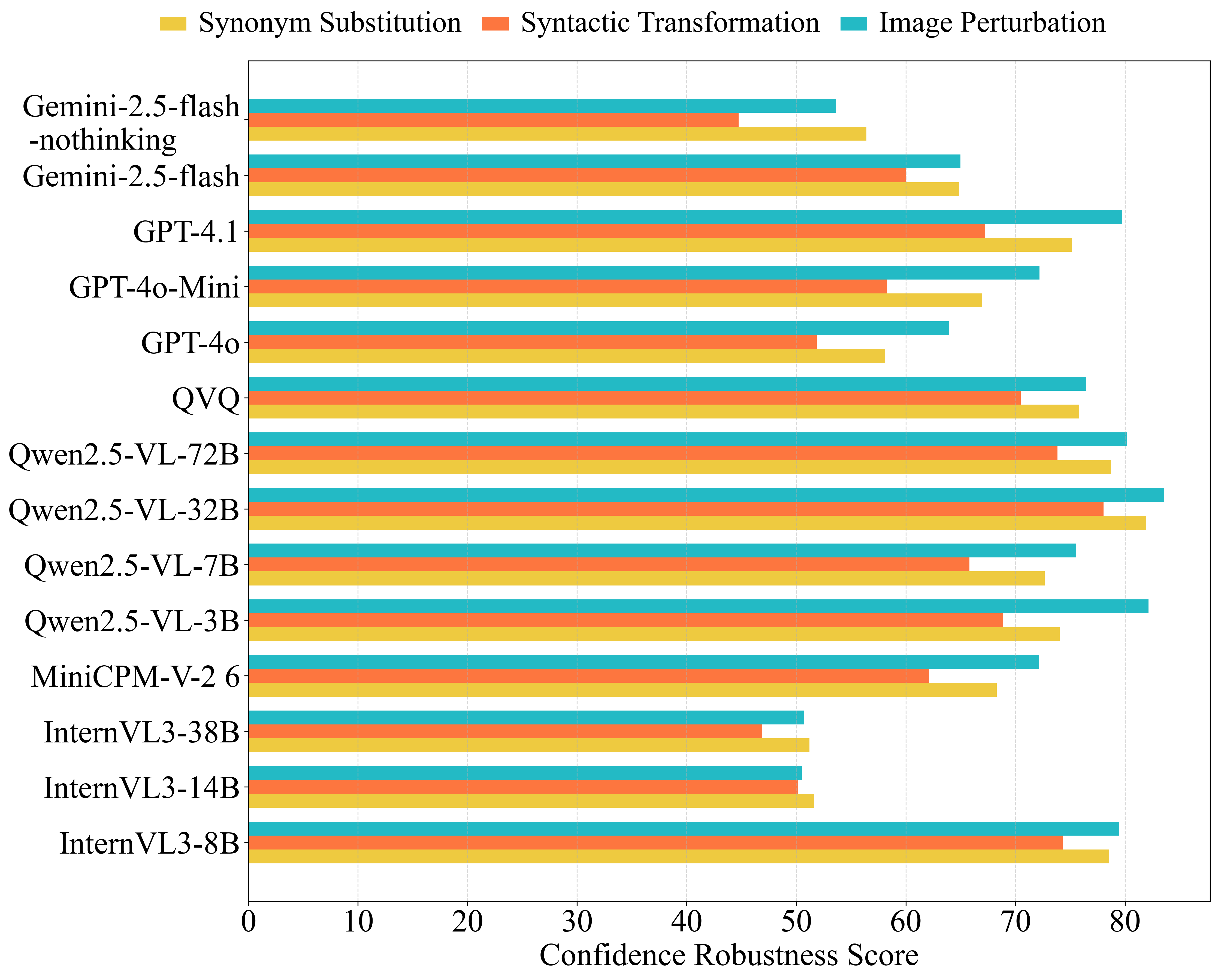}
  \caption{Confidence Robustness Score (CRS) under Different Perturbations}
  \label{fig:CSS_classify_3_horizontal}
\end{figure}

As shown in Figure~\ref{fig:CSS_classify_3_horizontal}, among all types of adversarial perturbations, MPJs exhibit the lowest confidence robustness scores (CRS) under syntactic transformations. This suggests that MPJs are least robust when facing syntactic transformations but semantically equivalent inputs. In contrast, they demonstrate stronger confidence robustness under synonym substitution and image perturbation. These results indicate that MPJs face considerable challenges in maintaining confidence robustness under syntactic transformations, while other types of adversarial perturbations also present non-negligible effects. Designing targeted strategies to enhance the confidence robustness of MPJs is crucial for obtaining reliable confidence estimates.

\section{Confidence Metrics across Difficulty Levels, Subjects, and Modalities}
\subsection{Confidence Metric Analysis across Different Difficulty Levels.}
\begin{figure*}[!htbp]
  \centering
  \includegraphics[width=0.9\linewidth]{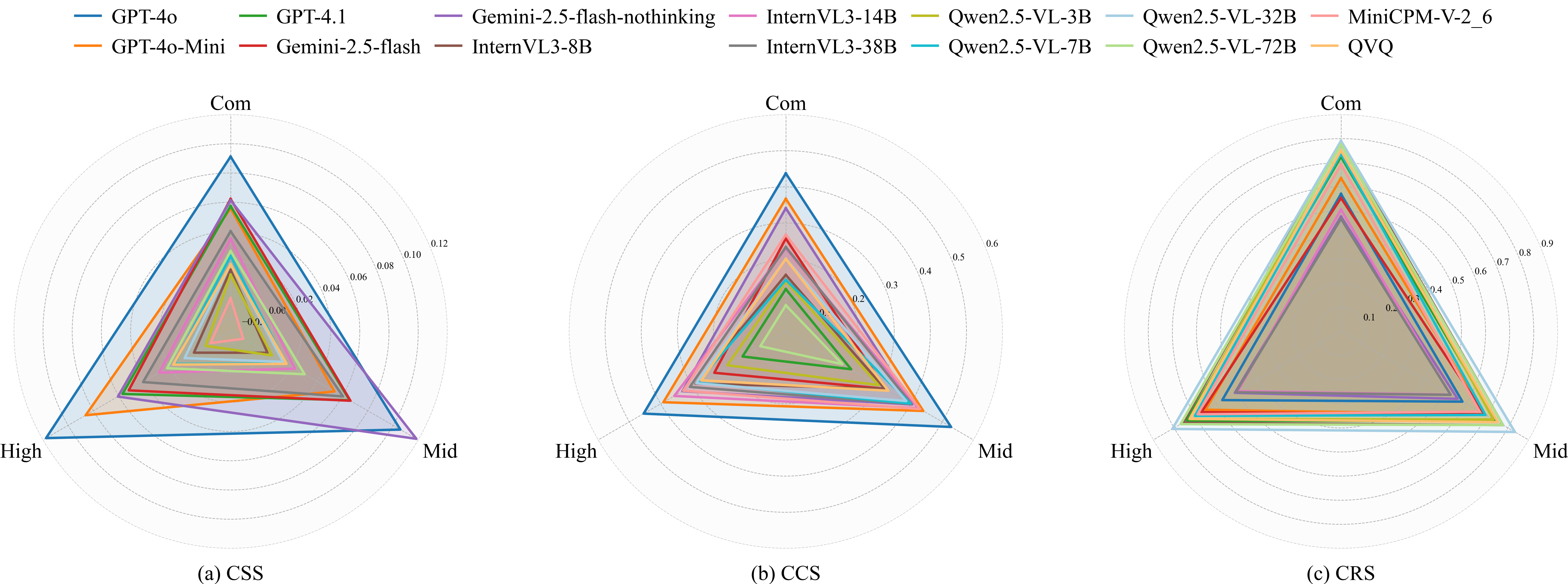}
  \caption{Confidence metric performance of MPJs across different difficulty levels.}
  \label{fig:difficulty}
\end{figure*}

The scores of the three core confidence metrics at different difficulty levels are shown in Figure~\ref{fig:difficulty}.
Most MPJs exhibit the highest CSS at the Middle School (Mid) level, with noticeable declines at High School (High) and Competition (Com) levels, though the trend is not strictly monotonic. In contrast, CCS shows a clear and consistent downward trend as difficulty increases, indicating that MPJs become increasingly miscalibrated, assigning overly high confidence to incorrect answers or low confidence to correct ones on harder problems. CRS, however, remains relatively stable across all difficulty levels for most MPJs, suggesting that confidence robustness to adversarial perturbations is not significantly affected by task complexity. These results reveal that while MPJs’ sensitivity and calibration degrade under more complex reasoning, their robustness remains largely unaffected, highlighting distinct challenges in improving confidence reliability across different dimensions.

\subsection{Confidence Metric Analysis across Different Input Modalities.}
\begin{figure*}[!htbp]
  \centering
  \includegraphics[width=0.9\linewidth]{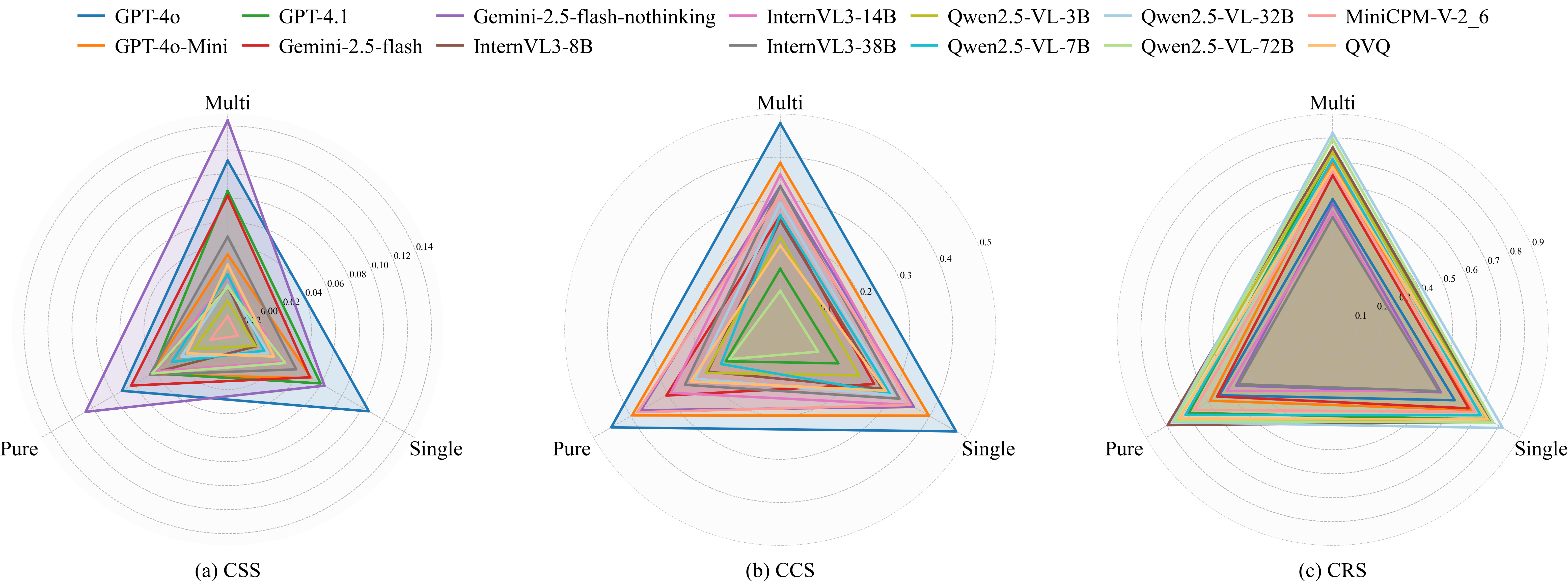}
  \caption{Confidence metric performance of MPJs across different input modalities.}
  \label{fig:modalities}
\end{figure*}

The scores of the three core confidence metrics across different input modalities are shown in Figure~\ref{fig:modalities}.
CSS shows clear modality dependence: most MPJs achieve higher scores in the Multi-image (Multi) setting than in Single-image (Single) or Pure-text (Pure), indicating that richer visual context enhances sensitivity to prediction correctness. In contrast, CCS remains largely consistent across modalities for most MPJs, suggesting limited influence of input type on calibration. Similarly, CRS scores are highly stable across modalities, indicating that robustness to perturbations is generally unaffected. Overall, input modality notably influences sensitivity, while calibration and robustness remain largely modality-invariant.

\subsection{Confidence Metric Analysis across Different Subject Domains.}
\begin{figure*}[!htbp]
  \centering
  \includegraphics[width=0.9\linewidth]{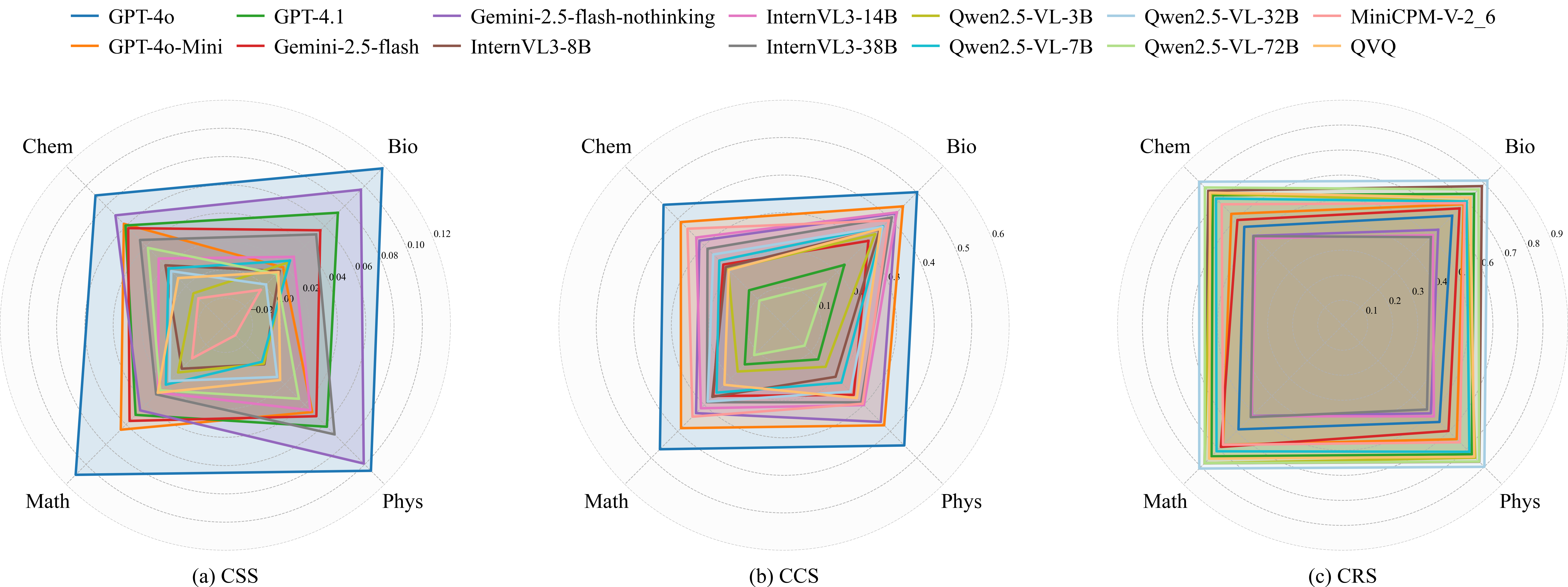}
  \caption{Confidence metric performance of MPJs across different subject domains.}
  \label{fig:subject}
\end{figure*}

The scores of the three core confidence metrics across different subject domains are shown in Figure~\ref{fig:subject}.
The performance of different MPJs on the Confidence Sensitivity Score (CSS) varies across subjects, but no consistent subject-specific trend is observed. This suggests that CSS is more dependent on model-specific characteristics rather than being driven by subject domain, implying that each MPJ may possess unique strengths and weaknesses when handling different types of knowledge structures or symbolic reasoning. Most MPJs achieve higher Confidence Calibration Scores (CCS) in the Biology domain, indicating better alignment between confidence and prediction correctness in that subject. In contrast, Confidence Robustness Scores (CRS) remain highly consistent across all subjects and MPJs, with radar plots forming near-square shapes, suggesting that subject domain has minimal impact on robustness. Overall, MPJs maintain consistent robustness against perturbations across tasks from different subject domains.

\section{High classification performance does not ensure confidence reliability.} 
As shown in Table~\ref{tab:main_table_with_F1}, strong classification performance of MPJs does not necessarily imply high confidence reliability. For instance, GPT-4o achieves a solid Macro F1 score of 78.12, indicating strong classification ability, yet its confidence sensitivity (CSS = 30.71) and calibration (CCS = 62.00) remain moderate. Similarly, while Gemini-2.5-flash attains the highest Macro F1 (81.74), its CCS (48.62) and robustness (CRS = 63.08) are not the best, revealing a mismatch between classification accuracy and confidence reliability. In contrast, GPT-4.1 demonstrates a more balanced profile, combining a high Macro F1 (80.87) with strong robustness (CRS = 73.62) and sensitivity (CSS = 38.51), though its CCS is relatively lower (37.65).

\begin{table*}[t]
\centering
\begin{tabular}{l r}
\toprule
\textbf{Statistic} & \textbf{Number} \\
\midrule
Total Samples & 1200 \\
Synonym Replacement / Sentence structure / Image Perturbations & 400 / 400 / 400 \\
Middle School / High School / Competition & 400 / 400 / 400\\
Math / Physics / Chemistry / Biology  & 400 / 400 / 400 \\
Single Image / Multi Images / Pure Text & 823 / 162 /215 \\

\bottomrule
\end{tabular}
\caption{Statistics of ConfProBench}
\label{tab:Statistics of ConfProBench}
\end{table*}

\begin{table*}[!htbp]
\centering
\begin{tabular}{|p{0.95\textwidth}|}
\hline
You are a sentence structure rewriting assistant. Your task is to rewrite a given sentence while altering its structure, ensuring that the original meaning is preserved. For each sentence, you must generate five distinct rewritten versions, each applying only one syntactic transformation. The goal is to create varied sentence structures while maintaining semantic accuracy and natural grammar. \\
\textbf{Syntactic Transformations (Choose One per Rewrite):} \\
Voice Change (Active $\leftrightarrow$ Passive)\\
2. Adverbial Position Adjustment \\
3. Clause Order or Structure Change \\
4. Phrase Structure Simplification or Expansion \\
5. Inversion or Emphatic Structure \\
6. Conditional / Purpose / Result Structure Transformation \\
\textbf{Key Constraints:} \\
- Preserve all steps in multi-step logical reasoning chains. \\
- Do not omit any mathematical derivations, steps, or intermediate expressions. \\
- Do not change numbers or mathematical expressions, including LaTeX formulas. \\
- Preserve meaning, grammar, and naturalness. \\
- Try to keep the length of the rewritten sentence close to the original (within 2–3 words difference). Avoid significant shortening or lengthening unless necessary for syntactic transformation. \\
- Only one syntactic transformation type per rewritten sentence. \\
\textbf{Output Format:} \\
\texttt{\{} \\
\quad "Original Sentence": "The original sentence", \\
\quad "Rewritten Sentences": [ \\
\qquad "rewritten sentence 1", \\
\qquad "rewritten sentence 2", \\
\qquad "rewritten sentence 3", \\
\qquad "rewritten sentence 4", \\
\qquad "rewritten sentence 5" \\
\quad ] \\
\} \\
\texttt{\# Student's solution: step-by-step student's solution} \\
\hline
\end{tabular}
\caption{Prompt for generating reasoning steps with syntactic transformation perturbations.}
\label{tab:syntactic_transformation_prompt}
\end{table*}

\begin{table*}[!htbp]
\centering
\begin{tabular}{|p{0.95\textwidth}|}
\hline
\textbf{Task Description:} You are a synonym substitution assistant. Given an input sentence, your task is to generate five distinct rewrites. In each version, you must replace at least one non-technical term with an appropriate synonym, and should replace as many non-technical terms as possible. Use different combinations of synonyms while keeping the original sentence structure and meaning intact. All outputs must be grammatically correct and sound natural. \\[0.8em]

\textbf{Definition:} Technical terms refer to specialized vocabulary that is specific to a particular field or discipline and should remain unchanged. These include, but are not limited to: mathematical symbols, scientific terminology, programming syntax, technical jargon, and domain-specific abbreviations. \\[0.8em]

\textbf{Key Constraints:} \\
- Do not modify any structural elements. \\
- Do not alter any numbers, numerical values, or mathematical expressions, including both plain numbers and LaTeX formulas. \\
- Do not change list symbols, bullet points, or any other sequence markers. \\
- Replace only the natural language content—do not alter formatting, technical terms, or domain-specific vocabulary. \\
- Ensure all rewritten sentences are grammatically correct, natural, and maintain the original meaning. \\
- Each rewritten version must replace at least one non-technical word, and should replace as many non-technical words as reasonably possible. \\[0.8em]

\textbf{Output Format:} \\
Provide your output in the following JSON structure: \\
\{\ \\
\quad "Original Sentence": "The original sentence", \\
\quad "Synonym Substitutions": [ \\
\quad\quad "Synonym Substitution 1", \\
\quad\quad "Synonym Substitution 2", \\
\quad\quad "Synonym Substitution 3", \\
\quad\quad "Synonym Substitution 4", \\
\quad\quad "Synonym Substitution 5" \\
\quad ] \\
\} \\[0.8em]

\texttt{\# Student's solution: step-by-step student's solution} \\
\hline
\end{tabular}
\caption{Prompt for generating reasoning steps with synonym substitution perturbations.}
\label{tab:synonym_substitution_prompt}
\end{table*}

\begin{table*}[!htbp]
\centering
\begin{tabular}{|p{0.95\textwidth}|}
\hline
You are a teacher skilled in evaluating the intermediate steps of a student's solution to a given problem. You are given a scientific problem, its correct final answer, and a student's step-by-step solution.\\

Your task is as follows:\\
1. Carefully solve the problem yourself, using the correct final answer as a hint to guide you to a consistent, correct reasoning path.\\
2. Then, evaluate the correctness of each step in the student's solution.\\

\textbf{For each step, output:}\\
- The full original step (as a string)\\
- A correctness label: \\
\quad - 1: if the model believes the step is correct (i.e., if P(correct)~$\geq$~0.5)\\
\quad - 0: otherwise\\
- A probability estimate P(correct) $\in$ (0, 1), representing the model's assessment of the likelihood that the step is correct (correctness label = 1)\\
- If the step is incorrect (correctness label = 0), also provide:\\
\quad - An error category (from the list below):\\

\quad\quad- Numerical Calculation Error\\
\quad\quad- Symbolic Calculation Error\\
\quad\quad- Visual Interpretation Error\\
\quad\quad- Reasoning Error\\
\quad\quad- Knowledge Error\\
\quad\quad- Question Understanding Error\\
\quad\quad- No solution provided\\

\textbf{Output Format:}\\
Wrap your output in this Python list format (and nothing else), enclosed by \textless evaluation\textgreater\ and \textless/evaluation\textgreater\ tags:\\

\texttt{<evaluation>} \\
\texttt{[} \\
\texttt{["Step 1: ...", correctness\_label, P\_correct, "Error type if incorrect"],} \\
\texttt{...} \\
\texttt{]} \\
\texttt{</evaluation>} \\

\textbf{Requirements:}\\
- You must return one and only one evaluation entry per step in the student's solution.\\
- The number of output entries must exactly match the number of steps (e.g., if the student has 15 steps, your output list must contain 15 entries).\\
- Do not skip, merge, or summarize steps.\\
- If the step is correct, use an empty string for the error type: \texttt{""}.\\
- Keep each step as a single complete unit, even if it contains multiple sentences.\\
- Please evaluate each step one by one. Every step must be assessed and scored individually, even if it is very short. Do not merge, omit, or skip any steps.\\
- Focus exclusively on the scientific, logical, or mathematical correctness of the solution. Ignore differences in formatting, expression style, specific wording, or presentation order, as long as the reasoning and results are valid. \\

\texttt{\# The given problem: \{problem\}} \\
\texttt{\# The Correct Final Answer: \{final answer\}} \\
\texttt{\# Student's solution: step-by-step student's solution} \\

\hline

\end{tabular}
\caption{Prompt for multimodal process judging.}
\label{tab:process_judging_prompt}
\end{table*}

\begin{table*}[htbp]
\centering
\begin{tabular}{lccccc}
\toprule
\textbf{Model} & \textbf{CRS$\uparrow$} & \textbf{CSS$\uparrow$} & \textbf{CCS$\uparrow$} & \textbf{Avg.$\uparrow$} & \textbf{Macro F1$\uparrow$}\\ 
\midrule

\multicolumn{5}{c}{\textbf{Open-source MLLMs}} \\ \midrule
InternVL3-8B           & 77.41 & 11.55 & 25.97   & 38.31 & 59.21\\
InternVL3-14B          & 50.78 & \underline{21.19} & \textbf{46.75}   & 39.57 & \underline{70.17}\\
InternVL3-38B          & 49.92 & \textbf{30.62} & \underline{44.49}   & \underline{41.68} & \textbf{73.66}\\
MiniCPM-V-2 6          & 68.05 & 6.60  & -47.95  & 8.90  & 38.31\\
Qwen2.5-VL-3B          & 74.71 & 3.15  & 2.73    & 26.86 & 50.90\\
Qwen2.5-VL-7B          & 71.19 & 10.38 & 15.80   & 32.46 & 56.88\\
Qwen2.5-VL-32B         & \textbf{81.06} & 15.93 & 41.60   & \textbf{46.20} & 67.13\\
Qwen2.5-VL-72B         & \underline{77.45} & 19.93 & 25.30   & 40.89 & 68.33\\
QVQ                   & 74.17 & 12.60 & 30.69   & 39.15 & 57.29\\
\midrule

\multicolumn{5}{c}{\textbf{Proprietary MLLMs}} \\ \midrule
GPT-4o                 & 57.37 & 30.71 & \textbf{62.00}   & \underline{50.03} & 78.12\\
GPT-4o-Mini            & \underline{65.58} & 13.03 & 47.73            & 42.11 & 66.08\\
GPT-4.1                & \textbf{73.62} & 38.51 & 37.65   & 49.93 & \underline{80.87}\\
Gemini-2.5-flash       & 63.08 & \textbf{48.29} & 48.62   & \textbf{53.33} & \textbf{81.74}\\
\makecell[l]{Gemini-2.5-flash\\-nothinking} & 51.20 & \underline{42.13} & \underline{51.55} & 48.29 & 79.02\\  

\bottomrule
\end{tabular}
\caption{Performance comparison across different MLLM-based Process
Judges on ConfProBench. The best performance for each metric is shown in bold, while the second-best is underlined.}
\label{tab:main_table_with_F1}
\end{table*}

\end{document}